\documentclass[11pt,a4paper]{article}
\usepackage[hyperref]{emnlp-ijcnlp-2019}
\usepackage{times}
\usepackage{latexsym}
\usepackage{comment}
\usepackage{color}
\usepackage{paralist}
\usepackage{xspace}
\usepackage{amsmath}
\usepackage{amssymb}
\usepackage{pgfplots}
\pgfplotsset{compat=newest}
\usepackage{multirow}
\usepackage{hyperref}
\usepackage{caption}
\aclfinalcopy 

\title{Controversy in Context}

\author{
	Benjamin Sznajder$^{\dagger}$
	\qquad Ariel Gera$^{\dagger}$ \qquad Yonatan Bilu$^{\dagger}$ \qquad Dafna Sheinwald$^{\dagger}$ \\ 
	\textbf{Ella Rabinovich$^{\dagger\blacktriangle}$ \qquad Ranit Aharonov$^{\dagger}$ \qquad David Konopnicki$^{\dagger}$ \qquad Noam Slonim$^{\dagger}$} \vspace{0.1cm} \\
	$^{\dagger}$IBM Research AI\\
	$^{\blacktriangle}$ Present affiliation: Dept. of Computer Science, University of Toronto \vspace{0.1cm} \\
	\{benjams, arielge, yonatanb, dafna, ranita, davidko, noams\}@il.ibm.com \\
	ella@cs.toronto.edu
}

\date{}

\begin{document}
\maketitle

\begin{abstract}
With the growing interest in social applications of Natural Language Processing and Computational Argumentation, 
a natural question is
how {\it controversial\/} a given concept is. 
Prior works relied on Wikipedia's metadata and on content analysis of the articles pertaining to a concept in question.
Here we show that the 
immediate {\em textual context} of a concept is strongly indicative of this property, 
and, using simple and language-independent machine-learning tools, we 
leverage this observation to achieve state-of-the-art results in controversiality prediction. 
In addition, we analyze and make available a new dataset of concepts labeled for controversiality. It is significantly larger than existing datasets, and grades concepts on a 0-10 scale, rather than treating controversiality as a binary label.  
\end{abstract}

\section{Introduction}
\label{intro}
Indicating that a web page is {\em controversial}, or disputed - for example, in a search result -
facilitates an educated consumption of the information therein, suggesting the content may not represent the ``full picture''. 
Here, we consider the problem of estimating the level of {\em controversiality} associated with a given  Wikipedia concept (title). We demonstrate that the textual contexts in which the concept is referenced can be leveraged to facilitate this.  

The definition of which concepts are {\em controversial} is controversial by itself; an accurate definition of this elusive notion 
attracted the attention of researchers from various fields, see for example some recent attempts in \cite{kittur2007he, shiriSIGIR2016, polishElsevier2018}. 

Most people would agree, for example, that {\em Global warming} is a controversial concept, whereas {\em Summer} is not. However, the concept {\em Pollution} may be seen as neutral by some, yet controversial by others, who associate it with environmental debates\footnote{These examples are based on the annotation of Dataset III in Section \ref{dataset}.}. In other words,
different people may have different opinions, potentially driven by different contexts salient in their mind. Yet, as reported in the sequel, an appreciable level of agreement can be reached, even without explicit context. 

Focusing here on Wikipedia concepts, 
we adopt as an initial ``ground truth'' the titles listed on the {\em Wikipedia list of controversial issues}\footnote{\href{https://en.wikipedia.org/wiki/Wikipedia:List\_of\_controversial\_ issues}{List\_of\_Controversial\_issues}}, 
which is curated based on so-called ``edit wars''. We then manually annotate a set of Wikipedia titles which are locked for editing, and evaluate our system on this much larger and more challenging dataset. 

To estimate the level of controversy associated with a Wikipedia concept, we propose to simply examine the words in the sentences in which the concept is referenced. Because a concept can often be found in multiple contexts, the estimation can be seen as reflecting the ``general opinion'' about it in the corpus.
This contrasts previous works, which consider this a binary problem, and employ a complex combination of features extracted from Wikipedia's article contents and inter-references, and more extensively -- from the rich edit history thereof. 

\section{Related work}
\label{related}
Analysis of controversy in Wikipedia, online news and social media has attracted considerable attention in recent years. 
Exploiting the collaborative structure of Wikipedia, estimators of the level of controversy in a Wikipedia article were developed based on the edit-history of the article \citep{kittur2007he, yasseri2012dynamics}. Along these lines, \citet{rad2012identifying} detect controversy based on mutual reverts, bi-polarity in the collaboration network, and mutually-reinforced scores for editors and articles. 
Similarly, 
\citet{shiriSIGIR2016} classify whether a Wikipedia page is controversial through the combined evaluation of the topically neighboring set of pages.

\emph{Content analysis} of controversial Wikipedia articles has been used to evaluate the level of controversy of other documents (e.g., web pages) by mapping them to related Wikipedia articles \citep{dori2013detecting}.
\citet{jang2016probabilistic} further 
build a {\em language model}, which enhances predictions made by existing classifiers, by inferring word probabilities from Wikipedia articles prominent in {\em Wikipedia controversy features} (mainly signals in edit history as discussed above) 
and from articles retrieved by manually selected query terms, believed to indicate controversy. 

\citet{choi2010identifying} detect controversy in news items by inspecting terms with excessive frequency in contexts containing sentiment words, and
\citet{sree2015controversy} study controversy in user comments of news articles using lexicons. 
Finally, \citet{jang2017modeling} suggest that controversy is not a universal but rather a community-related concept, and, therefore, should be studied in context.

Here we
measure a concept's controversiality from the explicit sentence-level context in which it is mentioned. In this, our approach is reminiscent of \citet{abstractness}, who suggest a similar approach to detect abstract concepts.

\section{Estimating a concept's controversiality level}
\label{predicting}
\subsection{Datasets}
\label{dataset}
We consider three datasets,
two of which are a contribution of this work.

{\em Dataset I} consists of 480 concepts previously analyzed in \citet{shiriSIGIR2016,rad2012identifying}. 
240 are positive examples, titles from the {\em Wikipedia list of controversial issues}, and 240 are negative examples chosen at random and exclusive of the positives. Over this dataset, we compare the methodology suggested here to those reported by  \citet{shiriSIGIR2016,rad2012identifying}. As the latter report overall accuracy of their binary prediction,
we convert our controversiality estimates
to a binary classification by 
classifying the higher-scored half 
as {\em controversial}, and the lower half as {\em non-controversial}.

{\em Dataset II} is based on a more recent version of the {\em Wikipedia list of controversial issues} (May 2017). As positive examples we take, 
from this list, 
all concepts\footnote{Excluding names of people, organizations, places, and artifacts; the same procedure was done for negative examples and for the concepts in dataset III.} which appear more than 50 times in Wikipedia. This leaves 608 controversial Wikipedia concepts. For negative examples, we follow \citet{shiriSIGIR2016,rad2012identifying} and select a like number of concepts at random\footnote{To verify that they were indeed non controversial, we annotated 200 of them.
96\% were indeed marked as non-controversial by a majority of the 10 required annotators.}. 
Here too, since each concept only has a binary label, we convert our estimation into a binary classification, and report accuracy.  

{\em Dataset III} is extracted from 3561 concepts whose Wikipedia pages are under {\em edit protection}\footnote{\href{https://en.wikipedia.org/wiki/Wikipedia:Lists\_of \_protected\_pages}{List\_of\_Protected\_Pages}}, assuming that many of them are likely to be controversial. They were then crowd-annotated, with 10 or more annotators per concept. 
The annotation instructions were: ``Given a topic and its description on Wikipedia, mark if this is a topic that people are likely to argue about.''.
Average pairwise kappa agreement on this task was 0.532. Annotations were normalized to controversiality scores on an integer scale of 0 - 10.\footnote{Datasets II and III are available for download  at \href{https://www.research.ibm.com/haifa/dept/vst/debating\_data.shtml}{Datasets}.}
We used this dataset 
for testing 
the models trained on Dataset I.

In all datasets, to obtain the sentence-level context of the concepts (positive and negative), we randomly select two equal-sized sets of Wikipedia sentences\footnote{Of length 10 to 70 tokens, as in \citet{abstractness}.}, that explicitly reference these concepts -- i.e., that contain a hyperlink to the article titled by the concept. 
Importantly, in each sentence we mask the words that reference the concept -- i.e., the surface form of the hyperlink leading to the concept -- by a fixed, singular token,
thus focusing solely on the context within which the concepts are mentioned.

\subsection{Controversiality Estimators}
We employ three estimation schemes based on the textual contexts of concepts.
The first relies on the context via pre-trained 
word embeddings of the concepts, which, in turn, 
are derived from the concepts' distributional properties in large samples of free texts. The other two schemes directly access the sentence-level contexts of the concepts.  
\\
\textbf{Nearest neighbors (NN) Estimator:}
We used the pre-trained GloVe embeddings \cite{glove} of concepts\footnote{The embedding of a multi-word concept was taken as the average of the embeddings of its individual words.} to implement a nearest-neighbor estimator as follows.
Given a concept $c$, we extract all labeled concepts within a given radius $r$ (cosine similarity $0.3$). In one variant, $c$'s controversiality score is taken to be the fraction of controversial concepts among them. 
In another variant, labeled concepts are weighted by their cosine similarity to $c$.
\\
\textbf{Naive Bayes (NB) Estimator:}
\label{NB}
A Naive Bayes model was learned, with a bag-of-words feature set, using the word counts in the sentences of our training data -- the contexts of the controversial and non-controversial concepts. The controversiality score of a concept $c$ for its occurrence in a sentence $s$, is taken as the posterior probability (according to the NB model) of
$s$ to contain a controversial concept, given the words of $s$ excluding $c$, and taking a prior of $0.5$ for controversiality (as is the case in the datasets). The controversiality score of $c$ is then defined as the average score over all sentences referencing $c$. 
\\
\textbf{Recurrent neural network (RNN):}
A bidirectional RNN using the architecture suggested in \citet{abstractness} was similarly trained. The network receives as input a concept and a referring sentence, and outputs a score. The controversiality score of a concept is defined, as above, to be the average of these scores. 


\subsection{Validation}
%
\subsubsection{Random $k$-fold}\label{random-k-fold}
We first examined the estimators in $k$-fold cross-validation scheme on the datasets I and II 
with $k=10$: the set of positive (controversial) concepts was split into 10 equal size sets, and the corresponding sentences were split accordingly.  Each set was matched with similarly sized sets of negative (non-controversial) concepts and corresponding sentences.  For each fold, a model was generated from the training sentences and used to score the test concepts. Scores were converted into a binary classification, as described in \ref{dataset}, and  accuracy was computed accordingly. Finally, the accuracy over the $k$ folds was averaged.

\subsubsection{Leave one category out}\label{thematic-vs-controversial}




In a preliminary task, we looked for words which may designate sentences associated with controversial concepts. To this end, we ranked the words appearing in positive sentences according to their {\em information gain} for this task. The top of the list comprises the following:
{\em that}, {\em sexual}, {\em people}, {\em movement}, 
{\em religious}, {\em issues}, {\em rights}.

The {\em Wikipedia list of controversial issues} specifies categories for the listed concepts, like Politics and economics, Religion, History, and Sexuality (some concepts are associated with two or more categories).  While some top-ranked words - {\em that, people, issues} - do seem to directly indicate controversiality \cite{ran:am2017,Roitman:2016}, others seem to have more to do with the category they belong to. Although these categories may indeed indicate controversiality, we consider this as an indirect or implicit indication, since it is more related to the controversial theme than to controversiality per-se.

To control for this effect, we performed a second experiment where we set the concepts from one category as the test set, 
and used the others for training 
(concepts associated with the excluded category are left out, regardless of whether they are also associated with one of the training categories). We did this for $5$ categories: History, Politics and economics, Religion, Science, and Sexuality. 
This way, thematic relatedness observed in the training set should have little or no effect on correctly estimating the level of controversy associated of concepts in the test set, and may even ``mislead'' the estimator.
We note that previous work on controversiality does not seem to address this issue, probably because the meta-data used is less sensitive to it.

\section{Results}
\label{results}
Table \ref{tab:accuracy} compares the accuracy reported on Dataset I for the methods  suggested in \citet{shiriSIGIR2016,rad2012identifying} with the accuracy obtained by our methods, as well as the latter on Dataset II, using $10$-fold cross-validation in all cases. Table \ref{tab:accuracy-cat} reports accuracy results of the more stringent analysis described in section \ref{thematic-vs-controversial}.


\begin{table}[h!]
	\begin{center}
    \small
		\begin{tabular}{|l|c|c|} 
			\hline
			\textbf{Classifier} & \textbf{Dataset} & \textbf{Acc} \\
			\hline
			\cite{rad2012identifying} (best)  &  
			 I & 0.84 \\
			\hline 
			\cite{rad2012identifying} 
            (practical) & 
			I & 0.75 \\
			\hline 
			\cite{shiriSIGIR2016} (best) &
			I & 0.74 \\
            \hline 
			Context based NN &
			I & 0.816 \\
            \hline 
			Context based NN (weighted) &
			I & 0.825 \\
			\hline
			Context based NB &
			I & 0.831 \\
			\hline
			Context based RNN  & 
			I & 0.865\\
			\hline
			\hline
			Context based NN  & 
			II & 0.775\\
            \hline 
			Context based NN (weighted) &
			II & 0.782 \\
			\hline
			Context based NB  & 
			II & 0.856\\
            \hline
			Context based RNN  & 
			II & 0.841\\
			\hline
		\end{tabular}
		\caption{Accuracy obtained by controversiality classifiers with $10$-fold cross validation.}
         \label{tab:accuracy}       
\vspace{1em}
    \small
		\begin{tabular}{|l|c|c|} 
			\hline
			\textbf{Classifier} & \textbf{Dataset} & \textbf{Acc} \\
			\hline
            Context based NN &  
			I & 0.788 \\
   			\hline
            Context based NN (weighted)& 
            I & 0.796 \\
			\hline
            Context based NB &  
			I & 0.813 \\
			\hline
			Context based RNN  &  
			I & 0.856 \\
			\hline
            \hline
            Context based NN &  
			II & 0.745 \\
   			\hline
            Context based NN (weighted)& 
            II & 0.763 \\
			\hline
            Context based NB &  
			II & 0.78 \\
			\hline
			Context based RNN  &  
			II & 0.784 \\
			\hline
		\end{tabular}
		\caption{Accuracy obtained by controversiality classifiers using leave-one-category-out cross validation.}
         \label{tab:accuracy-cat}       
\vspace{1em}
    \small
		\begin{tabular}{|l|c|c|} 
			\hline
            \textbf{Estimator}
			 &  \textbf{Corr} 
             & \textbf{Acc} \\
            \hline
            Context based NB &
			0.651 & 0.893\\
            \hline
            Context based RNN & 
			0.596 & 0.847\\
            \hline
            Context based BERT & 
			0.586 & 0.863\\
            \hline
	\end{tabular}
		\caption{Pearson Correlation and Accuracy obtained by using the models from Dataset I on Dataset III.}
         \label{tab:accuracy-III}       
       \end{center}
\end{table}

\citet{rad2012identifying} review several controversy classifiers. The most accurate one, the {\em Structure classifier}, builds, among others, collaboration networks by considering high-level
behavior of editors both in their individual forms, and
their pairwise interactions. A collaboration
profile containing these individual and pairwise features is
built for each two interacting editors and is classified based on the  agreement or disagreement relation between them. 
This intensive computation renders that classifier impractical. Table \ref{tab:accuracy} therefore also includes the most accurate classifier \citet{rad2012identifying} consider {\em practical}.

As seen in Table \ref{tab:accuracy}, for the usual $10$-fold 
analysis the simple classifiers suggested here are on par with the best and more 
complex classifier reported in \citet{rad2012identifying}. Moreover, in the leave-one-category-out setting (Table \ref{tab:accuracy-cat}), accuracy 
indeed drops, but only marginally.
We also observe the superiority of classifiers that directly access the context (NB and RNN) over classifiers that access it via word embedding (NN). 

Table \ref{tab:accuracy-III} presents results obtained when models trained on Dataset I are applied to Dataset III. For this experiment we also included a BERT network  \cite{devlin2018bert} fine tuned on Dataset I.
The Pearson correlation between the 
scores obtained via 
manual annotation and the scores generated by our automatic estimators suggests a rather strong linear relationship between the two. Accuracy was computed as for previous datasets, by taking here
as positive examples the concepts receiving 6 or more positive votes, and as negative a random sample of 670 concepts out of the 1182 concepts receiving no positive vote.





\section{Conclusions}
\label{conclusions}
We demonstrated that the 
sentence--level 
context in which a concept appears 
is indicative of its controversiality. This follows 
\citet{abstractness}, who show this for 
concept {\em abstractness} and suggest 
to 
explore further properties identifiable in this way. 
Importantly, we observed that this method may pick up signals which are not directly related to the property of interest. For example, since many controversial concepts have to do with religion, part of what this method may learn is thematic relatedness to religion. However, when controlling for this effect, the drop in accuracy is fairly small.


The major advantages of our estimation scheme are its simplicity and reliance 
on abundantly accessible features. At the same time, its accuracy is similar to state-of-the-art classifiers, which depend on complex meta-data, and rely on sophisticated - in some cases impractical - algorithmic techniques. Because the features herein are so simple, 
our estimators are convertible to any corpus, in any language, even of moderate size. 

Recently, IBM introduced {\em Project Debater} \cite{debater}, an AI system that debates humans on controversial topics. Training and evaluating such a system 
undoubtedly requires an extensive supply of such topics, which can be enabled by the automatic extraction methods 
suggested here as well as the new datasets. 


\section*{Acknowledgment}
We are grateful to Shiri Dori-Hacohen and Hoda Sepehri Rad for sharing their data with us and giving us permission to use it.
\bibliography{controversiality}
\bibliographystyle{acl_natbib_nourl}

\end{document}